\pdfoutput=1

\documentclass[11pt]{article}

\usepackage[preprint]{acl}

\usepackage{times}
\usepackage{latexsym}
\usepackage{listings}
\usepackage{pythonhighlight}
\usepackage{pythontex}
\usepackage{booktabs}
\usepackage{tabularx} 
\usepackage{fontawesome5}
\usepackage{hyperref}
\usepackage{multicol}
\usepackage{lipsum}
\usepackage{subcaption}
\usepackage{listings}

\usepackage{xcolor}
\usepackage{listings}
\usepackage[scaled=0.85]{inconsolata} 

\definecolor{keywordblue}{rgb}{0.26, 0.34, 0.75}
\definecolor{stringred}{rgb}{0.64, 0.11, 0.12}
\definecolor{graycomment}{rgb}{0.35, 0.45, 0.45}
\definecolor{numbercolor}{rgb}{0.1, 0.5, 0.5}

\lstdefinestyle{absynthstyle}{
  language=Python,
  basicstyle=\ttfamily\small,
  keywordstyle=\color{keywordblue}\bfseries,
  stringstyle=\color{stringred},
  commentstyle=\color{graycomment}\itshape,
  numberstyle=\color{numbercolor},
  identifierstyle=\color{black},
  showstringspaces=false,
  frame=single,
  breaklines=true,
  tabsize=4
}

\makeatletter
\newcommand{\github}[1]{%
   \href{#1}{\faGithubSquare}%
}
\usepackage[T1]{fontenc}

\usepackage[utf8]{inputenc}

\usepackage{microtype}

\usepackage{inconsolata}

\usepackage{graphicx}

\title{TRACE: Training and Inference-Time Interpretability Analysis for Language Models}

\author{Nura Aljaafari$^{1\dagger}$,~ Danilo S. Carvalho$^{3}$,~ Andr\'{e} Freitas$^{1,2,3}$ \\
  $^{1}$ Department of Computer Science, University of Manchester, United Kingdom\\
  $^{2}$ Idiap Research Institute, Switzerland\\
  $^{3}$ National Biomarker Centre, CRUK-MI, Univ. of Manchester, United Kingdom\\
  \texttt{\{firstname.lastname\}@[postgrad.]$^{\dagger}$manchester.ac.uk}\\
  \href{https://github.com/neuro-symbolic-ai/trace_package}{
    \includegraphics[height=2.5ex]{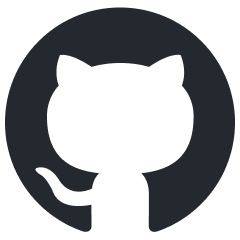}
    \hspace{0.3em}\texttt{github.com/neuro-symbolic-ai/trace\_package}
  }
  \href{https://youtu.be/8UxNH3_xpno}{
    \includegraphics[height=2.5ex]{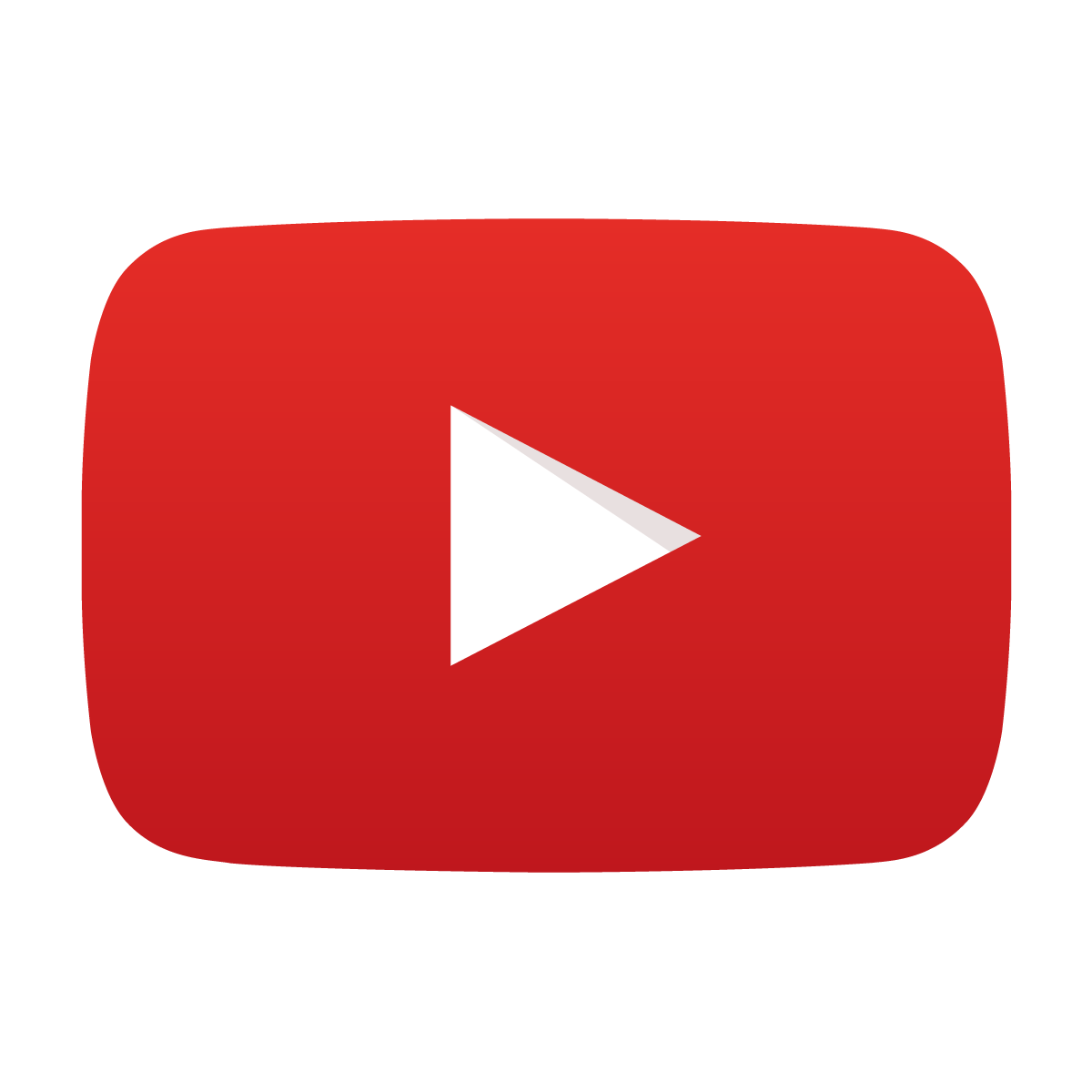}
    \hspace{0.3em}\texttt{Short Video}
  }
  }

\begin{document}
\maketitle
\begin{abstract}
Understanding when and how linguistic knowledge emerges during language model training remains a central challenge for interpretability. Most existing tools are post hoc, rely on scalar metrics, or require nontrivial integration effort, making comprehensive interpretability analysis difficult to deploy and maintain. We introduce \textbf{TRACE}, a modular toolkit for training and inference-time interpretability analysis of transformer models. It enables lightweight, in-training analysis of linguistic and representational signals, including features probing, intrinsic dimensionality, Hessian curvature, and output diagnostics. It integrates with \textbf{ABSynth}, a controllable synthetic corpus generator that provides structured annotations for precise evaluation of linguistic feature acquisition. Experiments with autoregressive transformers demonstrate that TRACE reveals developmental phenomena such as early syntactic emergence, delayed semantic acquisition, and representational compression, signals overlooked by traditional scalar metrics such as loss or accuracy. With minimal integration effort, the tool enables layer-wise diagnostics, convergence-based early stopping, and detection of structural errors, making transformer analysis interpretable, actionable, and reproducible.

\end{abstract}

\section{Introduction}\label{sec:intro}

Interpreting the behaviour of transformer-based language models (LMs) has been gaining interest and importance \cite{nostalgebraist-2020, wang2023interpretability, hanna2023how,belrose2023eliciting, meng2022locating}, especially with the widespread use of them across various domains. This interpretation is fundamental for verifying their correctness, analysing reasoning processes, and improving robustness \cite{bereska2024mechanistic}.

Although various approaches to interpretability have been proposed, most are post hoc: they examine fully trained models using input-output correlations \cite{kokhlikyan2020captum, lundberg2017unified, zhao2024reagent}, structured tasks such as mathematics or algorithmic reasoning \cite{hanna2023how, nanda2023progress}, or mechanistic analysis \cite{belrose2023eliciting, meng2022locating}.
\begin{figure*}[h]
    \centering
    \includegraphics[width=\linewidth]{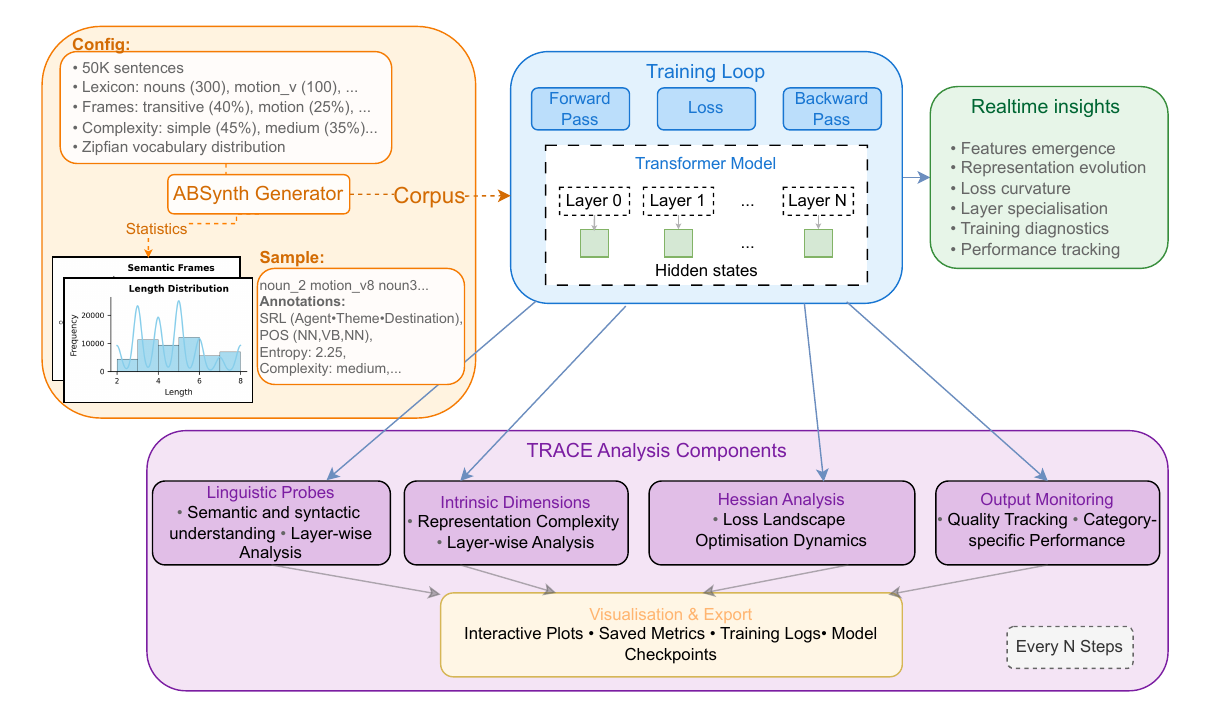}
    \caption{TRACE system overview. ABSynth generates controlled linguistic datasets with explicit annotations, which feed into transformer training loops instrumented with lightweight analysis hooks. Four modular components track linguistic emergence: semantic/syntactic probes, intrinsic dimensionality analysis, Hessian landscape exploration, and output monitoring. All modules generate automated visualisations and can operate independently during training or post-hoc analysis.}
    \label{fig:overview}
\end{figure*}
Tools that operate in training time, such as TensorBoard or Weights \& Biases, focus narrowly on scalar metrics like loss or gradient norms. They reveal \textit{whether} a model is learning, but not \textit{what} is being learned or \textit{when} key capabilities emerge. Lightweight, modular tools for tracking representational development remain scarce.

This creates a critical gap: researchers lack accessible tools for comprehensive analysis of \textit{when} and \textit{how} semantic structure emerges, whether during training or in deployed models. As a result, the relationship between optimisation objectives and representational development remains poorly understood, despite its importance for training decisions, generalisation, and debugging failures. This blind spot hinders both scientific understanding of language acquisition in LMs and the practical optimisation of model training workflows.

We introduce \textbf{TRACE} (Tracking Representation Abstraction and Compositional Emergence), a modular toolkit for training and inference-time interpretability analysis of transformer models.\footnote{TRACE \& ABSynth are released under GPLv3 License. \label{fn:trace_license}} TRACE exposes internal learning dynamics through lightweight instrumentation requiring minimal code changes, enabling structured tracking of semantic emergence, representational compression, and loss landscape evolution—signals that are invisible to conventional training logs.

To support systematic experimentation, TRACE pairs with \textbf{ABSynth}, a controllable synthetic corpus generator that provides structured linguistic data with aligned annotations\footref{fn:trace_license}. While TRACE can analyse any transformer training process, ABSynth enables precise, repeatable experiments grounded in known linguistic structures.

TRACE is designed for researchers and developers seeking a modular, easy-to-integrate interpretability tool to better understand, debug, or intervene on transformer models. Demonstrations on decoder-only models trained with ABSynth reveal distinct developmental phases in syntax and semantics, compression of internal representations, and curvature shifts in optimisation dynamics, insights not available through traditional monitoring alone. Thus, it transforms training from an opaque optimisation process into an interpretable, research-ready workflow for studying representational learning and guiding practical model development, with additional support for inference-time analysis.

\section{Overview}\label{sec:overview}
\textbf{TRACE} (Figure~\ref{fig:overview}) is a modular toolkit for interpretability analysis of transformer training. It supports dynamic monitoring of linguistic, geometric, and optimisation signals, integrating seamlessly into training loops and model evaluation pipelines. It is also model-agnostic, extensible, and requires only minimal code changes. By default, it supports the construction of standard transformer architectures as described in \cite{vaswani2017attention}, and adapting to custom models typically requires only minor adjustments. A companion dependency, ABSynth, supplies linguistically annotated training data with controllable complexity.

\subsection{Design Overview}
The toolkit consists of two core modules and one external dependency:
\begin{itemize}
    \item \textbf{Monitoring Hooks}: Injected into the training loop or inference pipelines, these capture hidden states, gradients, and loss signals without interfering with model execution.

    \item \textbf{Analysis Modules}: Pluggable components for probing, intrinsic dimensionality estimation, loss analysis, and output-level diagnostics. 

    \item \textbf{ABSynth (external dependency)}: A synthetic corpus generator that provides structured, annotated training data with controllable linguistic complexity.

\end{itemize}

\subsection{Key Capabilities}
TRACE combines four capabilities that are difficult to achieve simultaneously within existing interpretability frameworks:
\begin{itemize}
    \item \textbf{Minimal Integration Effort:} Requires only a few lines of code to enable comprehensive analysis, with modular components that can be added, removed, or extended without code refactoring.
    
    \item \textbf{Flexible Analysis Scope:} Supports both live analysis during training and post-hoc inspection of trained models, continuously monitoring linguistic probes, representational geometry, and loss curvature.
    
    \item \textbf{Automated Diagnostics:} Automatically plots learning curves, phase changes, and divergence indicators, with outputs saved in structured formats (e.g. JSON, CSV) for further analysis.
    
    \item \textbf{Actionable Insights:} Reveals dynamics informing early stopping, architecture tuning, and training schedule adjustments. Custom probes and datasets are registered through a simple interface.
\end{itemize}

\section{Controlled Corpus Generation (ABSynth)}\label{sec:abysnth}
ABSynth is a synthetic corpus generator that produces linguistically annotated datasets with explicitly controlled syntactic and semantic structures. Inspired by frame semantics \cite{baker1998berkeley, fillmore1982frame}, ABSynth models language as a composition of semantic frames, participant roles, and lexical units. Unlike natural corpora, which entangle linguistic properties or lack annotations, ABSynth provides fine-grained control over linguistic dimensions while generating English-like data. This enables systematic experiments on representational learning, generalisation, and abstraction.

\subsection*{Minimal API and Basic Usage}
ABSynth supports rapid corpus generation with a simple interface:


\begin{lstlisting}
from absynth.corpus import SyntheticCorpusGenerator
generator = SyntheticCorpusGenerator()
corpus = generator(10000)
corpus.save("corpus_full.json", indent=2)
\end{lstlisting}

A typical generated sentence appears as:
\begin{quote}
\texttt{noun139 transitive\_verb8s noun40 preposition4 location2}
\end{quote}

This sentence instantiates the \texttt{transitive\_action} semantic frame with structured annotations automatically generated during corpus creation.

\subsection*{Sentence Structure and Generation}
Each sentence is constructed from a semantic frame containing a syntactic template. Frames specify roles (e.g., Agent, Patient, Location), which are mapped to lexical items from Zipfian-distributed pools \cite{zipf1949human, piantadosi2014zipf}. Templates determine constituent order (e.g., \texttt{[arg0, verb, arg1, prep, arg2]}) and syntactic structure. ABSynth automatically generates token-level annotations, including part-of-speech tags, semantic roles with positions, and metadata as follows:

\begin{itemize}
    \item \textbf{Semantic Roles:}
    \begin{itemize}
        \item \texttt{noun139} — Agent (position 0)
        \item \texttt{noun40} — Patient (position 2)  
        \item \texttt{location2} — Location (position 4)
    \end{itemize}
    \item \textbf{POS Tags:} \texttt{[NN, VB, NN, IN, NN]}
    \item \textbf{Metadata:} Complexity = \texttt{medium}; Entropy = 2.25; Length = 5
\end{itemize}
Corpus properties can be customised via frame and complexity distributions:
\begin{lstlisting}
corpus = generator.generate_corpus(
    num_sentences=10000,
    complexity_distribution={
        "simple": 0.55, "medium": 0.35, 
        "complex": 0.1},
    semantic_frame_distribution={
        "transitive_action": 0.4,
        "motion": 0.3,
        "intransitive_action": 0.3})
\end{lstlisting}
Key controllable dimensions include: (i) \textbf{Lexical selection:} Tokens sampled from role-specific pools with Zipfian \citep{zipf1949human, piantadosi2014zipf} distributions; (ii) \textbf{Frame selection and creation:} Controls argument structure and semantic relations; (iii) \textbf{Complexity distribution:} Specifies proportions of simple, medium, and complex constructions; and (iv) \textbf{Statistical properties:} Entropy profiles, collocational strengths, and predictability patterns.


\section{Dynamic Linguistic Tracking (TRACE)}\label{sec:tool_sample_usage}
As a sample use case, we present results from training a 2-layer decoder-only transformer with \texttt{3} attention heads, \texttt{384}-dimensional MLP, and \texttt{96}-dimensional hidden states on an ABSynth-generated corpus of \texttt{50K} examples (batch size \texttt{128}, learning rate \texttt{$1e-3$}, \texttt{70K} steps). All analysis modules were enabled: probing, intrinsic dimensionality, Hessian curvature, and output diagnostics. While we show a combined diagnostic plot for conciseness, each module also outputs standalone visualisations and structured logs. A complete walk-through example of the tool usage, including detailed inputs, outputs, and interpretations, is provided in Appendix~\ref{sec:usecase}. 


\subsection*{API Integration}
TRACE wraps a standard training loop with minimal configuration:

\begin{lstlisting}
from trace.training import Trainer, TrainingConfig
config = TrainingConfig(
    epochs=10,
    track_interval=500,                
    save_visualization=True 
)
trainer = Trainer(config, tokenizer, model)
trainer.train(train_loader, val_loader)
\end{lstlisting}
All modules run independently, log structured outputs (e.g., JSON, CSV), and support training and post-hoc visualisation.

\begin{figure*}[h]
    \centering
    \begin{minipage}[t]{0.49\linewidth}
        \centering
        \includegraphics[width=\linewidth]{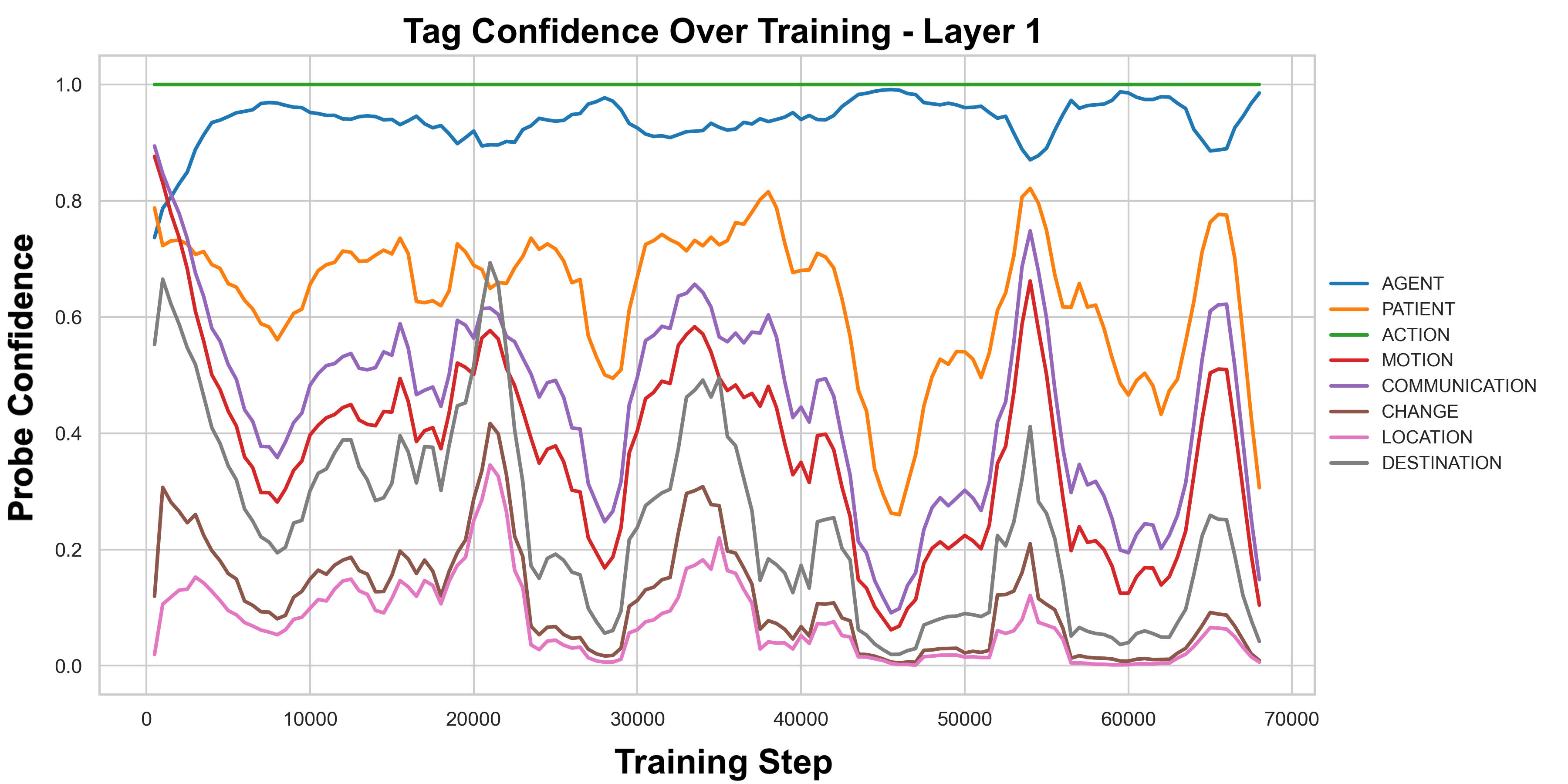}
        \caption*{\small (a) Probe confidence over semantic roles}
    \end{minipage}
    \hfill
    \begin{minipage}[t]{0.48\linewidth}
        \centering
        \includegraphics[width=\linewidth]{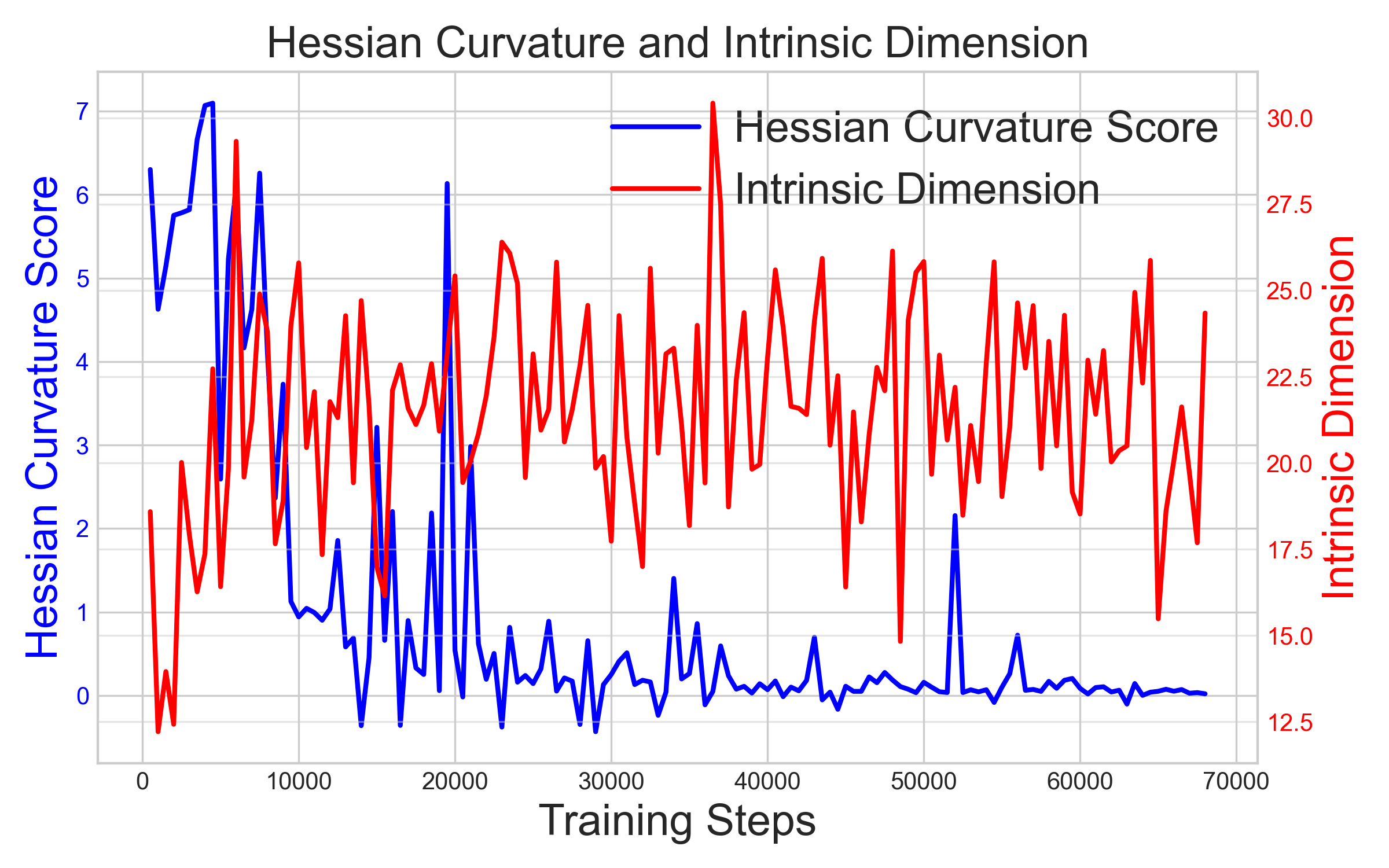}
        \caption*{\small (b) Intrinsic dimension and curvature evolution}
    \end{minipage}
    \caption{\textbf{Sample TRACE visual diagnostics.} (a) Semantic probe confidence per role at decoder layer 1. TRACE tracks dynamic shifts, revealing phase transitions in linguistic encoding (e.g., mid-training dips and late recovery). (b) Joint evolution of Hessian curvature and average intrinsic dimensionality, exposing compression, abstraction, and optimisation phase shifts. Together, these metrics offer temporal insights missed by scalar loss alone.}
    \label{fig:trace_visuals}
\end{figure*}

\paragraph{Linguistic Structure Probing.} Semantic and syntactic probes can be applied to any layer, supporting both static and dynamically updated models. Probes return confidence per linguistic role, revealing where and when linguistic features emerge and consolidate. 
\begin{lstlisting}
config = TrainingConfig(
    track_semantic_probes=True,
    probe_load_paths={ 
        (0, 'decoder'): './probes/pos_layer0.pt',
    }
)
\end{lstlisting}

Figure~\ref{fig:trace_visuals}a shows semantic probe confidence for the decoder layer 1. Core roles (e.g., \texttt{AGENT}, \texttt{ACTION}) stabilise early, while adjunct roles (e.g., \texttt{LOCATION}, \texttt{DESTINATION}) fluctuate, indicating delayed consolidation. Several dips in confidence align with changes in curvature and intrinsic dimensionality (Figure~\ref{fig:trace_visuals}b), suggesting linked representational shifts. The key capabilities of this module include:

\begin{itemize}
    \item \textbf{Multi-label probing:} Simultaneous detection of multiple linguistic features
    \item \textbf{Category-specific analysis:} Performance breakdown by syntactical and semantic categories
    \item \textbf{Emergence tracking:} Identification of critical learning phases for different linguistic structures
\end{itemize}

\paragraph{Intrinsic Dimensionality Analysis.} Representational complexity is estimated through intrinsic dimensionality (ID) metrics, including TwoNN \cite{facco2017estimating} and PCA-based estimators \cite{cangelosi2007component}:

\begin{lstlisting}
config = TrainingConfig(
    track_intrinsic_dimensions=True,          
    id_method="TwoNN")
\end{lstlisting}

ID can be reported per layer or model average, capturing when and where representation compression or expansion occurs across model layers. 
Figure~\ref{fig:trace_visuals}b shows an early drop in average dimensionality followed by a sharp rebound and stabilisation. These trends suggest a shift from early overcompression to stabilised abstraction. Such transitions coincide with early probe confidence dips, supporting the view that dimensionality evolution reflects changes in internal representational. These measures can be used to detect abstraction shifts, potential representational bottleneck or underutilised representational capacity. 

\paragraph{Optimisation Landscape Analysis.} 

Loss curvature is tracked using Lanczos-based Hessian approximations \cite{lanczos1950iteration}. The framework captures gradient–Hessian alignment, dominant eigenvalue shifts, trace, and other spectral metrics:

\begin{lstlisting}
config = TrainingConfig(
    hessian_n_components=10,             
    track_component_hessian=True,
    track_gradient_alignment=True)
\end{lstlisting}

In Figure~\ref{fig:trace_visuals}b, early spikes in curvature precede shifts in ID, revealing a transition from sharp loss regions to flatter ones, memorisation and generalisation. Additional moderate spikes later in training suggest ongoing adjustments during fine-tuning phases. Beyond this visualisation, the tool simultaneously monitors other optimisation metrics, including: 

\begin{itemize}
    \item \textbf{Curvature evolution:} Changes in loss landscape sharpness and flatness, trace, and dominant eigenvalues.
    \item \textbf{Component analysis:} Separate tracking of components (e.g. attention, feed-forward) contributions
    \item \textbf{Memorisation detection:} Train/validation landscape divergence analysis
    \item \textbf{Gradient alignment:} Relationship between optimisation direction and principal curvature
\end{itemize}

\paragraph{Output-Level Diagnostics.} Model predictions are analysed by token class and semantic role, enabling stratified accuracy reporting across linguistic categories. This diagnostic helps detect linguistic structural misalignment, where the model predicts the correct type (e.g., a noun or a patient role) but fails to recover the intended lexical item (e.g., \texttt{noun3} vs. \texttt{noun542}). Such errors indicate representational drift: the model learns the general grammatical form but lacks alignment to specific semantic slots or referents.

\begin{lstlisting}
config = TrainingConfig(
    track_semantic_roles=True,       
    semantic_roles_granularity='detailed' )
\end{lstlisting}

\paragraph{System Evaluation.}
This demonstration serves as a functional evaluation of TRACE. The example training run reveals distinct, interpretable dynamics across modules, semantic emergence in probe confidence, representational compression through intrinsic dimensionality drops, and curvature shifts in the optimisation landscape. The observed alignment between independent metrics further validates the utility of TRACE for diagnosing learning phase transitions. Full training configuration is provided in Appendix~\ref{app:experimental}, and further comparisons with existing interpretability tools appear in Table~\ref{tab:baseline-comparison}.
\paragraph{Research Questions and Applications.} TRACE addresses fundamental questions in transformer interpretability: (i) \textbf{Emergence timing:} When and where do syntactic and semantic features first appear?; (ii) \textbf{Representational evolution:} Which layers compress or expand representational space during training?; (iii) \textbf{Generalisation patterns:} Is the model developing structural understanding or overfitting to surface patterns?; and (iv) \textbf{Training efficiency:} Can representation convergence inform early stopping decisions?

\paragraph{Automatic Visualisation and Reporting.} All modules generate comprehensive visualisations automatically, including: (i) \textbf{Linguistic confidence evolution:} Per-category confidence scores across training steps; (ii) \textbf{Dimensionality trajectories:} Layer-wise ID evolution with compression detection; (iii) \textbf{Hessian landscape analysis:} Eigenvalue evolution and curvature dynamics; and (iv) \textbf{Performance breakdown:} Category-specific accuracy trends and structural alignment metrics. Results are also saved in standard formats (e.g., CSV, JSON) for further analysis and integration with external tools. By exposing model internals at different stages, TRACE makes transformer training interpretable, inspectable, and diagnostic by design.
\begin{table*}[h]
\centering
\small
\begin{tabularx}{\textwidth}{lXXXX}
\toprule
\textbf{Feature} & \textbf{TRACE (Ours)} & \textbf{Post-hoc Probing} & \textbf{Manual PCA / ID} & \textbf{TransformerLens} \\
\midrule
\textbf{Timing of Analysis}   & During and after training & After training & After training & During and after training \\
\textbf{Layer-wise Tracking}  & Yes & Yes & Yes & Yes \\
\textbf{Temporal Resolution}  & High (per N steps) & Sparse (checkpoints) & Sparse & Sparse (checkpoints) \\
\textbf{Model Support}        & Any custom PyTorch model & Any checkpoint & Any checkpoint & HuggingFace models only (limited) \\
\textbf{User Effort}          & Low (1 config file) & High (custom scripts) & High & High (custom pipelines) \\
\textbf{Causal Intervention Support}   & Partial (custom scripting required) & No & No & Partial (custom scripting required) \\
\textbf{Emergence Detection}  & Yes & No & No & No \\
\textbf{Training Support} & Yes (native) & No & No & Partial (custom scripting required) \\
\textbf{Supports Early Stopping}   & Yes (live signals) & No & No & No \\
\textbf{Gradient/Loss Monitoring}  & Yes & Indirect & N/A & Yes \\
\bottomrule
\end{tabularx}
\caption{Comparison of TRACE against post-hoc interpretability workflows and TransformerLens. TRACE provides low-effort, modular interpretability analysis with native training support, while existing tools require custom scripting and lack temporal tracking capabilities.}
\label{tab:baseline-comparison}
\end{table*}
\section{Related works}\label{sec:related_wrok}
\paragraph{Post-hoc Representational Analysis.}
A common approach to interpreting language models involves probing classifiers or sparse autoencoders (SAEs) trained on frozen representations. Probes evaluate the presence of linguistic features using lightweight supervised models \citep{hewitt2019structural, belinkov2018evaluating}, while SAEs attempt to recover features in latent spaces in an unsupervised setting \citep{bricken2023monosemanticity, kantamneni2025sparse}. While both techniques can be adapted for use during training, they are typically applied post hoc and require manual effort for their models construction, retraining, and integration. As standalone tools, they provide limited visibility into when features emerge or how they evolve.

\paragraph{Attribution and Causal Methods.}
Attribution tools (e.g., integrated gradients \citep{sundararajan2017axiomatic} and attention flow analysis \citep{voita-etal-2019-analyzing}) and causal interventions (e.g., activation patching \citep{meng2022locating} and path patching \citep{wang2023interpretability}) identify critical inputs or subcircuits. However, they operate on static checkpoints and provide no insight into the dynamics of representational development. 

\paragraph{Model Inspection Frameworks.}
Frameworks like BertViz \citep{vig-2019-multiscale}, InterpretDL \citep{JMLR:v23:21-0738}, and TransformerLens \citep{nanda2022transformerlens} enable weight and attention analysis. TransformerLens partially supports training-time instrumentation, but requires nontrivial modification. Platforms like TensorBoard or Weights \& Biases \citep{wandb} track scalar metrics, offering no view into linguistic or geometric structure. 

\paragraph{Controlled Datasets and Linguistic Benchmarks.} Several datasets evaluate LMs' capabilities. Static benchmarks such as GLUE \citep{wang2018glue}, BLiMP \citep{warstadt2020blimp}, and COGS \citep{kim-linzen-2020-cogs} evaluate linguistic competence but lack structural flexibility. CounterFact \citep{meng2022locating} and IOI \citep{wang2023interpretability} target specific phenomena, while mathematical datasets \citep{power2022grokking, saxton2019analysing} offer training control but limited linguistic coverage. ABSynth provides extensible, richly annotated data aligned with TRACE’s dynamic instrumentation. 

\paragraph{Comparison with Existing Tools.} Table~\ref{tab:baseline-comparison} highlights how TRACE differs from typical interpretability pipelines and frameworks. It supports native training-time analysis, live monitoring, and plug-and-play extensibility with structured logging.







\section{Conclusion}
We presented \textbf{TRACE}, a modular toolkit for interpretability of language model, and \textbf{ABSynth}, a companion corpus generator for controlled linguistic experimentation. TRACE enables in-training analysis of linguistic, geometric, and optimisation signals, shifting interpretability from a post-hoc task to a dynamic diagnostic process. Applied to a decoder transformer, TRACE revealed phase-structured training dynamics, including semantic emergence, representational compression, and curvature shifts, not visible through traditional training metrics. These signals also enable targeted interventions such as early stopping, architecture tuning, and training schedule adjustment. The tool’s modular design allows each component to operate independently or in combination, enabling a broad range of interpretability research. While current demonstrations use synthetic corpora, TRACE can generalise to natural language models. Planned extensions include Hugging Face integration and support for automatic annotation via tools such as NLTK, further broadening TRACE’s scope for real-world applications. By making internal learning dynamics observable and actionable, TRACE advances both theoretical insight and practical control over LM analysis.

\bibliography{custom}

\appendix
\section{Experimental setup}
\label{app:experimental}
\subsection{Software Environment and Dependencies}

All experiments were run on a single NVIDIA RTX A6000 GPU. The environment was built using Python 3.11.13 and configured via Conda. Core libraries include PyTorch (v2.7.1) and ABSynth (v0.1.1) for synthetic dataset generation (with the same requirements listed here). Other tools include scikit-learn (v1.7.0), SciPy (v1.15.3), NumPy, and scikit-dimension (v0.3.4). Visualisation components rely on Matplotlib (v3.10.3), Matplotlib-inline (v0.1.7), and Seaborn (v0.13.2). Additional packages include tqdm (v4.67.1). The full setup is specified in the environment.yml file included with the code repository and is fully reproducible.

\subsection{Model Architecture}

We use a lightweight Transformer architecture to enable rapid training and frequent checkpointing. This allows TRACE to track representational evolution with minimal overhead.

\begin{itemize}
    \item \textbf{Architecture:} Decoder-only Transformer
    \item \textbf{Layers:} 2 decoder layers
    \item \textbf{Hidden size:} 96
    \item \textbf{Feed-forward dimension:} 384
    \item \textbf{Attention heads:} 3
    \item \textbf{Maximum sequence length:} 16
\end{itemize}

\subsection{Training Configuration}

\begin{itemize}
    \item \textbf{Corpus:} 25,000 synthetic sentences
    \item \textbf{Batch size:} 128
    \item \textbf{Epochs:} 500
    \item \textbf{Learning rate:} $1\times10^{-4}$
    \item \textbf{Optimizer:} Adam
    \item \textbf{Tracking frequency:} Every 500 steps
\end{itemize}

\section*{Appendix C: Licensing}
TRACE is released under the GNU General Public License v3.0 (GPLv3). This license ensures that the software remains free and open-source. Users are free to use, modify, and distribute the code, provided that derivative works also adopt the GPLv3 license.

\newpage
\onecolumn

\section{Example use case}\label{sec:usecase}
This appendix illustrates TRACE's functionality and interpretability workflow. It includes an end-to-end case study using a transformer model trained on an ABSynth-generated corpus. This walkthrough demonstrates how TRACE can be used to monitor semantic emergence during training and interpret the resulting behaviour via multiple analytical lenses.

\subsection{ABSynth Data Generation}\label{sec:data_app}
We provide an example of data generation, where we set all the parameters, but \textbf{emphasise} that the user has the liberty to use the default configurations by specifying only the number of examples needed, as shown in Section \ref{sec:abysnth}.

\subsubsection{ABSynth Input Configuration}
\begin{lstlisting}
from absynth.lexicon import Vocabulary, LexiconGenerator

# Define vocabulary sizes
vocab = Vocabulary({
    "noun":300, "transitive_verb":40, "intransitive_verb":25, 
    "communication_verb":20, "motion_verb":20, "change_verb":15, "adjective":40, 
    "adverb":25, "location":150, "temporal":35, "instrument":25, "preposition":15, 
    "conjunction":10, "determiner":8
})

lexicon = LexiconGenerator(
    vocab_sizes=vocab,           # Custom vocabulary sizes
    num_clusters=5,              # Number of semantic clusters to create
    zipfian_alpha=1.05,             # Alpha parameter for Zipfian distribution
    error_bias=0.00001,              # Error bias for word generation
    random_seed=42               # For reproducible generation
)

from absynth.sentence import SentenceGenerator, FrameManager
templates = FrameManager()
sentence_generator = SentenceGenerator(lexicon, templates)

from absynth.corpus import SyntheticCorpusGenerator
generator = SyntheticCorpusGenerator(lexicon=lexicon, sentence_generator=sentence_generator)
corpus = generator.generate_corpus(
    num_sentences=25000,
    complexity_distribution={"simple": 0.55, "medium": 0.35, "complex": 0.10},
    semantic_frame_distribution={
        "transitive_action": 0.1,
        "transitive_with_location": 0.15,
        "motion_with_source": 0.15,
        "temporal_action": 0.15,
        "instrumental_action": 0.15,
        "multi_action": 0.15,
        "temporal_complex": 0.15,
    }
)
from absynth.visualization import Visualizer
visualizer = Visualizer(log_dir='./plots')
visualizer.visualize(corpus)
\end{lstlisting}
This setting creates a corpus of 25K sentences, with a mix of syntactic templates and semantic frames. Sentences vary in their surface structure but preserve underlying argument structures aligned with FrameNet-like role schemas.

\subsubsection{Sample Generated Data}
A representative sentence from the generated corpus is:
\begin{lstlisting}
Input: "noun139 transitive_verb8s noun40 preposition4 location2"
\end{lstlisting}
The corresponding annotations include both structural and semantic metadata:

\begin{lstlisting}
{
  "sentence": "noun139 transitive_verb8s noun40 preposition4 location2",
  "semantic_roles": {
    "noun139": {"role": "Agent", "position": 0},
    "noun40": {"role": "Patient", "position": 2},
    "location2": {"role": "Location", "position": 4}
  },
  "pos_tags": ["NN", "VB", "NN", "IN", "NN"],
  "metadata": {
    "complexity": "medium",
    "frame": "transitive_action",
    "length": 5,
    "entropy": 2.25
  }
}
\end{lstlisting}
This format enables fine-grained probing of semantic representations during training and evaluation. Each token is associated with a role and position, and every sentence is traceable to its underlying generation rule.
\subsubsection{Corpus Statistics and Visualisation}
\begin{figure}[ht]
\centering
\begin{subfigure}[t]{0.48\textwidth}
  \centering
  \includegraphics[width=\linewidth]{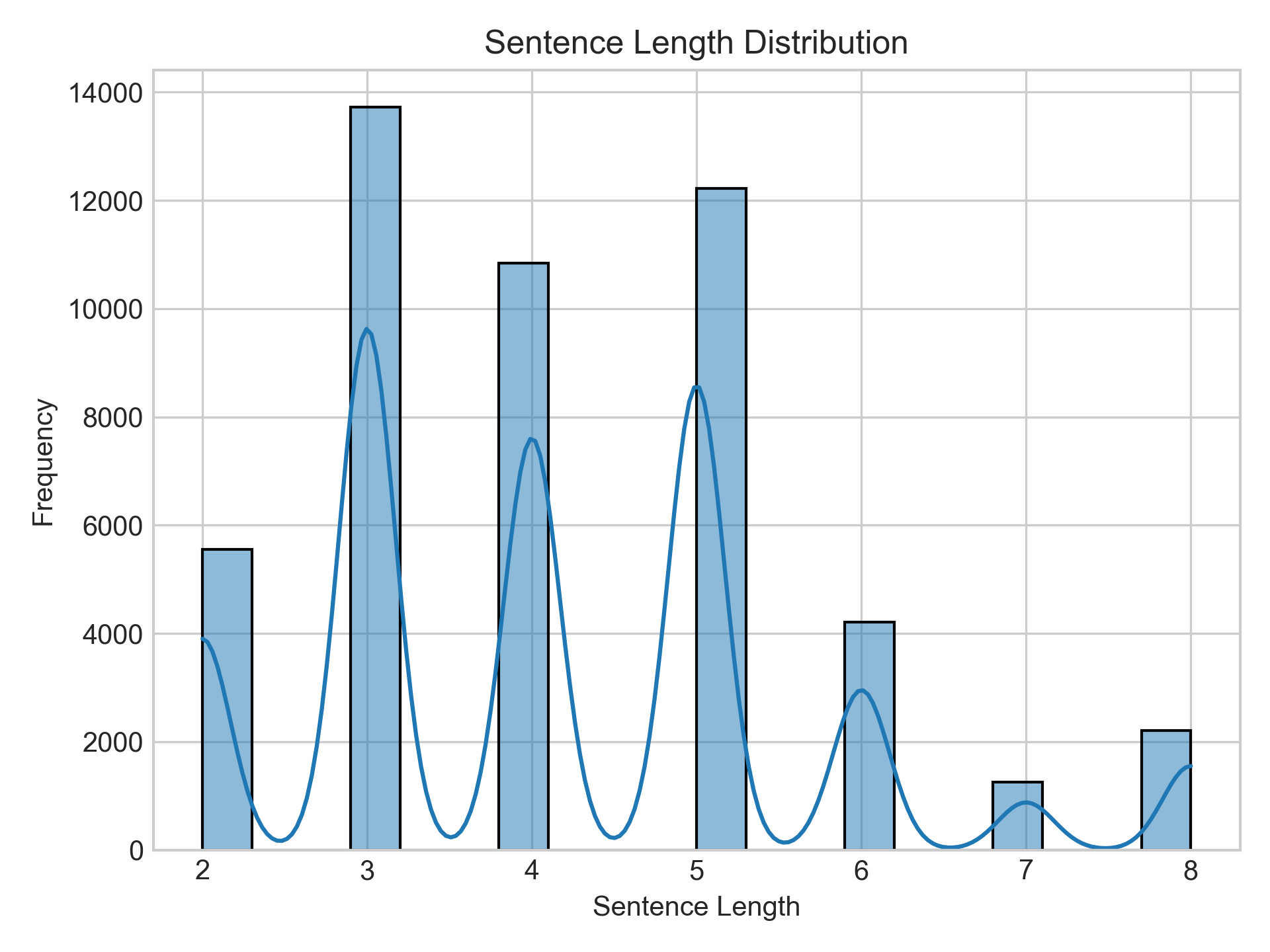}
  \caption{Hessian trace over training steps.}
  \label{fig:sentence_length_hist_kde}
\end{subfigure}
\hfill
\begin{subfigure}[t]{0.48\textwidth}
  \centering
  \includegraphics[width=\linewidth]{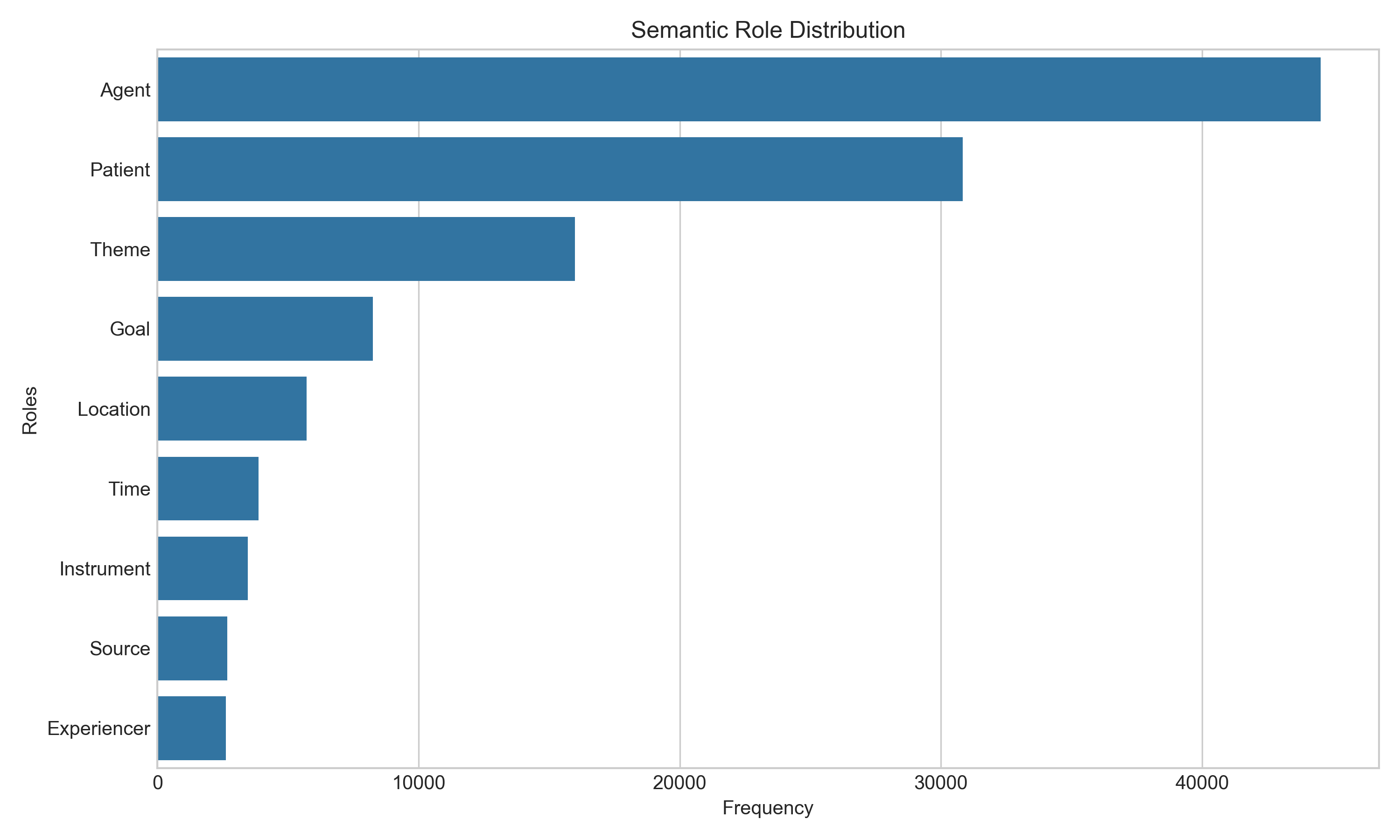}
  \caption{Gradient and Hessian norm dynamics.}
  \label{fig:semantic_role_distribution}
\end{subfigure}
\caption{(Sample statistical visualisation of ABSynth: (Left) Distribution of sentence lengths in the generated ABSynth corpus. (Right) Frequency distribution of semantic roles across the dataset.}
\label{fig:absynth_stats}
\end{figure}

\begin{table}[h]
\centering
\begin{tabular}{|l|c|}
\hline
\textbf{Metric} & \textbf{Value} \\
\hline
Total sentences & 25,000 \\
Vocabulary size & 7910 tokens \\
Average sentence length & 4.17012 tokens \\
Semantic frame coverage & 6 distinct frames \\
Zipfian compliance & $\alpha = 1.05$ \\
\hline
\end{tabular}
\caption{ABSynth corpus summary statistics.}
\label{tab:absynth_stats}
\end{table}

\subsection{Model Training Setup}\label{sec:training_app}
\subsubsection{Model Configuration}
To evaluate TRACE’s interpretability modules in a controlled training setting, we trained a lightweight decoder-only Transformer on the ABSynth corpus introduced in Section~\ref{sec:tool_sample_usage}. This setup enables fine-grained temporal analysis of representational and optimisation dynamics. While this demonstration focuses on decoder-only architectures, we note that TRACE also supports encoder-only and encoder-decoder models.
\begin{lstlisting}
model_config = TransformerConfig(
    model_type="decoder_only",
    vocab_size=7000,
    d_model=96,              # Hidden dimension
    num_heads=3,             # Attention heads  
    num_decoder_layers=2,    # Number of layers
    d_ff=384,               # Feed-forward dimension
    max_seq_length=16,      # Maximum sequence length
    dropout=0.1
)
\end{lstlisting}
\subsubsection{Training Configuration}
    
    
\begin{lstlisting}
training_config = TrainingConfig(
    epochs=30,
    learning_rate=1e-3,
    batch_size=128,
    
    # Enable all analysis modules
    track_hessian=True, # Loss landscape analysis
    track_linguistic_probes=True, # POS understanding
    track_semantic_probes=True, # Semantic role understanding
    track_intrinsic_dimensions=True, # Representation dimensionality
    track_pos_performance=True,  # Output POS accuracy
    track_semantic_roles_performance=True,  # Output semantic accuracy
    probe_load_paths=probe_paths,
    semantic_probe_load_path=semantic_probe_paths,
    
    # Analysis frequency
    track_interval=500,      # Every 500 steps
    save_visualization=True
)
\end{lstlisting}
\subsection{TRACE Analysis Results}
This section presents representative outputs from TRACE’s core modules, applied to the synthetic training run described in Sections \ref{sec:data_app}–\ref{sec:training_app}. These examples demonstrate how TRACE captures distinct learning dynamics across layers, linguistic abstractions, and training stages. All inputs and outputs are automatically logged by TRACE when the corresponding tracking module is enabled by the user. 

\subsubsection{Linguistic Probe Evolution}
We trained semantic role probes on model checkpoints throughout training to extract interpretable structure from hidden representations. The probes are trained to predict semantic roles and POS based on intermediate activations. This allows us to track when and where in training various linguistic categories emerge.

\paragraph{Input:} Hidden states $h$ from decoder layers 0 and 1, sampled at regular training intervals ($500$ in this experiment).
 
\begin{figure}[h]
\centering
\begin{minipage}{0.85\textwidth}
  \centering
  \includegraphics[width=\linewidth]{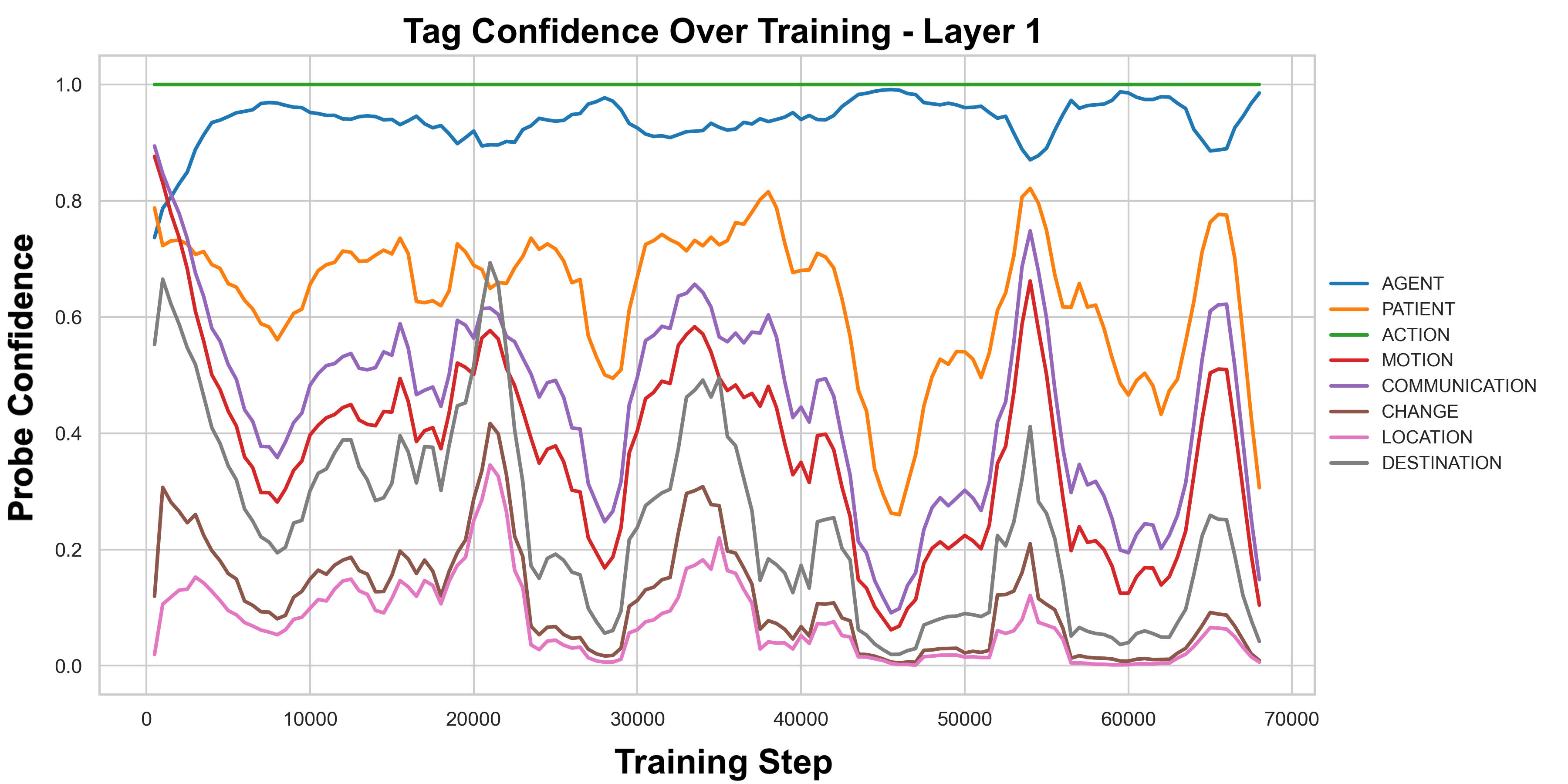}
  \captionof{figure}{Semantic role probe scores across training steps. Layer 1 shows alignment with core roles over time with steps where there was higher alignment and other for lower.}
  \label{fig:probe_plot_app}
\end{minipage}%
\hfill
\label{fig:probes_combined}
\end{figure}

\paragraph{Output Example: }
Figure~\ref{fig:probe_plot_app} shows probe confidence scores for Layer 1 across training steps. Confidence values indicate the strength of alignment between hidden states and semantic roles.

\paragraph{Interpretation:}
\begin{itemize}
     \item \textbf{Core Role Emergence:} AGENT, ACTION, and PATIENT show rapid gains in probe confidence, stabilising early in training. These roles are central to the predicate-argument structure, and their early emergence suggests the model internalises core semantics first.

    \item \textbf{Adjunct Role Delay:} LOCATION and other adjunct roles show lower and more delayed gains, indicating that peripheral semantic categories are acquired later and less robustly.

    \item \textbf{Fluctuation Signals Reorganisation:} Adjunct roles exhibit greater fluctuations in confidence across training steps. These fluctuations may reflect ongoing representational reorganisation as the model adjusts the alignment of semantic roles across layers and training stages.
        
    \item \textbf{Semantic Acquisition Hierarchy:} The emergence pattern suggests a learning hierarchy, with core argument structure learned before modifiers, consistent with theoretical linguistics and prior neural acquisition work.

    \item \textbf{Alignment with Representational Dynamics:} Confidence dips at intermediate steps (e.g., 10k–25k) align with known transitions in intrinsic dimensionality and Hessian curvature, reflecting internal restructuring as the model shifts from memorisation to abstraction.
\end{itemize}

\subsubsection{Intrinsic Dimensionality Analysis}
\paragraph{Input: } Hidden states $h$ from decoder layers 0 and 1, sampled at regular training intervals ($500$ in this experiment).

\begin{figure}[ht]
\centering
\begin{minipage}{0.50\textwidth}
  \centering
  \includegraphics[width=\linewidth]{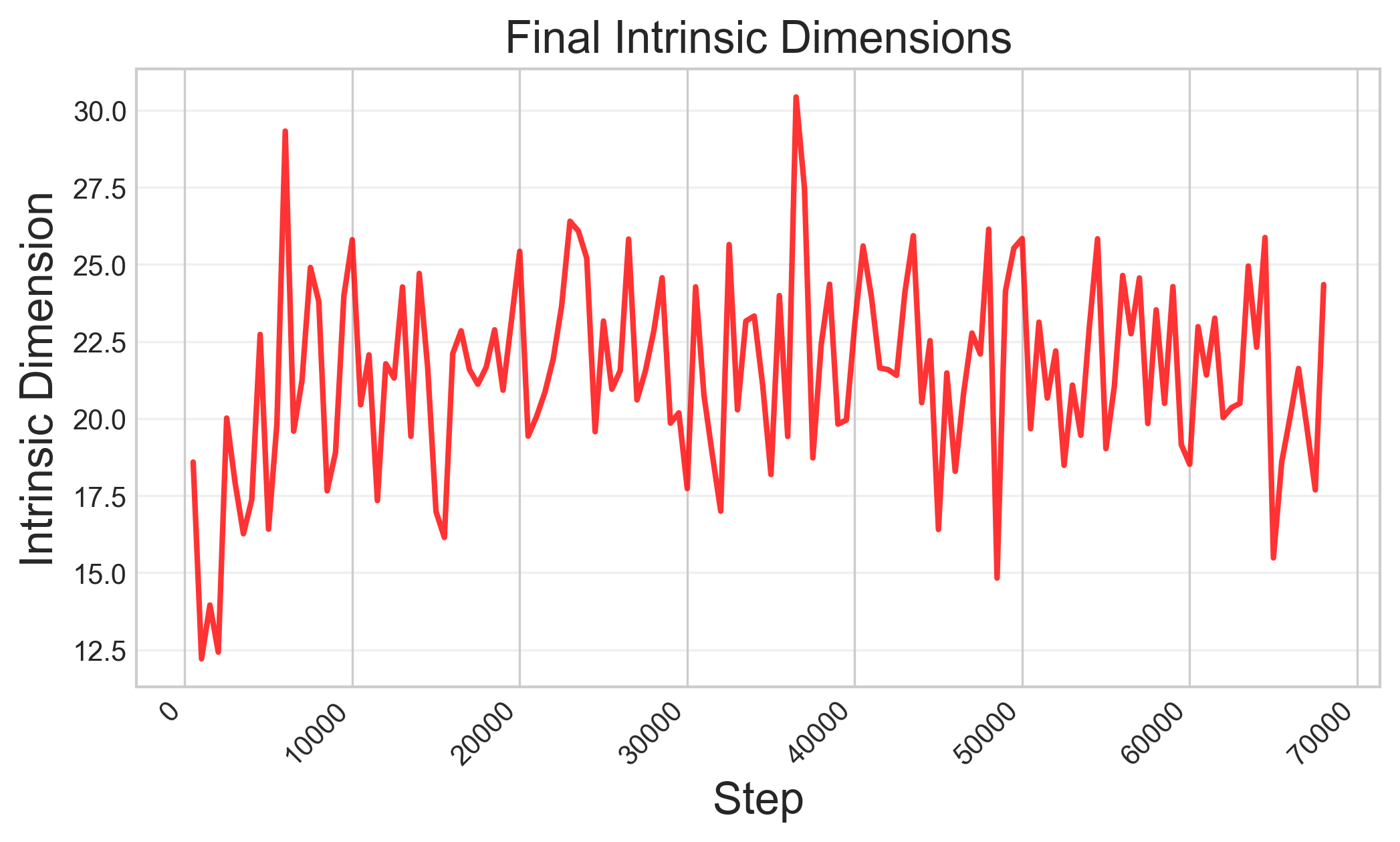}
  \captionof{figure}{Intrinsic dimensionality across training steps.}
  \label{fig:id_plot}
\end{minipage}%
\hfill
\begin{minipage}{0.46\textwidth}
  \centering
  \begin{tabular}{|l|c|}
  \hline
  \textbf{Training Step} & \textbf{Avg. ID} \\
  \hline
  500  & 18\\
  2,000  & 12\\
  5,000  & 16 \\
  10,000 & 25 \\
  25,000 & 23\\
  \hline
  \end{tabular}
  \captionof{table}{Sample averaged Intrinsic dimensionality over layers across training steps.}
  \label{tab:id_sample}
  \vspace{0.5em}
  
\end{minipage}
\end{figure}

\paragraph{Output Example: }
Figure \ref{fig:id_plot} and Table \ref{tab:id_sample} show a sample of the ID analysis output, the figure shows the overall ID over the whole course of training steps, while the table shows a sample of the results. 

\paragraph{Interpretation:}
\begin{itemize}
    \item \textbf{Initial Compression and Expansion Pattern:} Early and sharp drop, 18 to 12 (steps 500-2,000) followed by and expansion to 25, patterns indicating transitioning from simple pattern matching to more complex linguistic understanding. 
    \item \textbf{Dynamic Training Regime:} Persistent oscillations between ID~17-25 throughout training suggest active representational restructuring rather than static convergence, reflecting adaptive complexity based on linguistic content. 
    
    \item \textbf{Cross-Metric Transitions:} Notable jumps in intrinsic dimensionality coincide with decreases in Hessian trace and fluctuations in probe confidence. These temporal alignments suggest coordinated signals between them.
\end{itemize}
\subsubsection{Hessian Landscape Analysis}
\paragraph{Input:} Loss gradients and second-order information calculated by our tool. 

\begin{figure}[ht]
\centering
\begin{subfigure}[t]{0.48\textwidth}
  \centering
  \includegraphics[width=\linewidth]{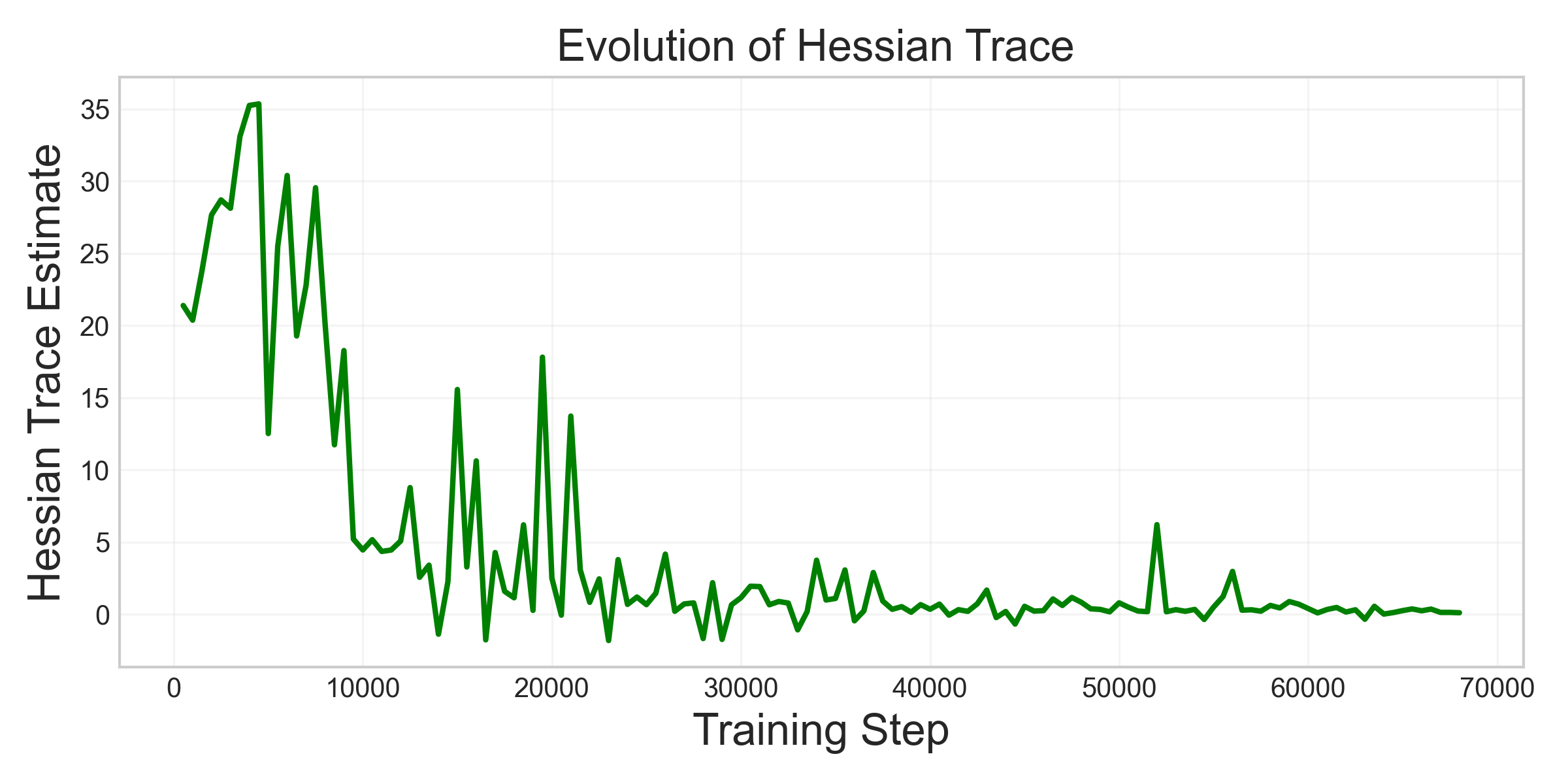}
  \caption{Hessian trace over training steps.}
  \label{fig:hessian_evolution}
\end{subfigure}
\hfill
\begin{subfigure}[t]{0.48\textwidth}
  \centering
  \includegraphics[width=\linewidth]{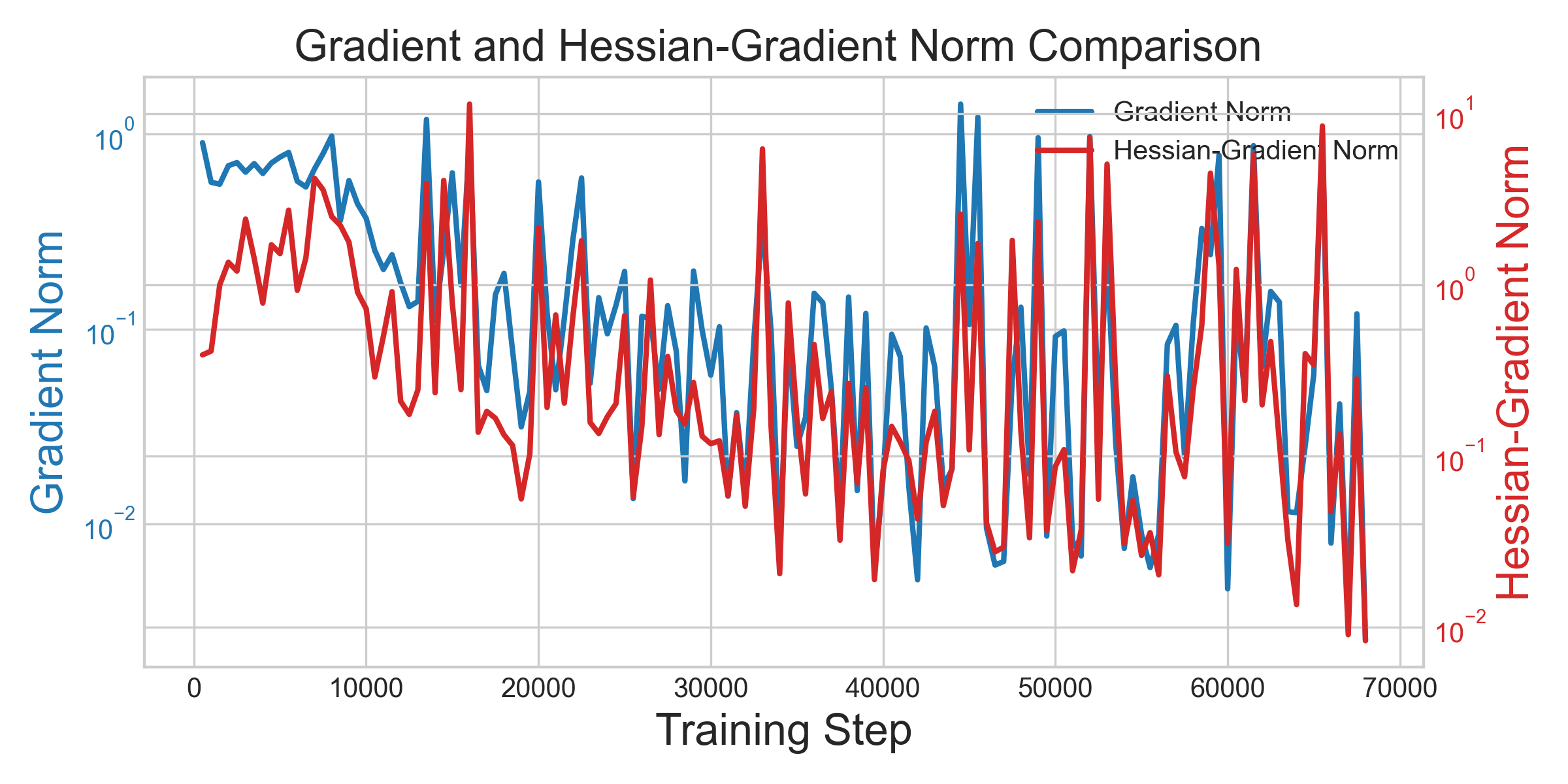}
  \caption{Gradient and Hessian norm dynamics.}
  \label{fig:sub_grad_norms}
\end{subfigure}
\caption{Evolution of curvature and norm characteristics during training.}
\label{fig:main_training_dynamics}
\end{figure}

\paragraph{Sample output:}
Figure~\ref{fig:main_training_dynamics} demonstrates the evolution of Hessian-based metrics throughout training, revealing characteristic patterns of curvature dynamics and spectral properties.

\paragraph{Interpretation:}
\begin{itemize}

        \item \textbf{Landscape Evolution Dynamics:} The Hessian trace exhibits a pronounced early peak followed by steady decrease, reflecting the model’s transition from highly curved regions associated with memorisation to flatter regions indicating abstraction and generalisation. This is common in randomly initialised networks.

        \item \textbf{Curvature–Dimensionality Duality:} An interesting pattern is observed between curvature and intrinsic dimensionality: periods of rising dimensionality coincide with drops in the Hessian trace, suggesting that representational expansion is facilitated by local flattening in the optimisation landscape.

        \item \textbf{Structural Transition Markers:} Isolated spikes in Hessian curvature, e.g.  around 55k steps, serve as indicators of local representational restructuring. These events temporally co-occur with inflection points in probing metrics and effective rank changes, underscoring the diagnostic value of second-order geometry for tracking internal phase shifts.

\end{itemize}
\subsubsection{Output Quality Monitoring}
\paragraph{Input: } Model predictions at regular training intervals, evaluated against gold-standard semantic role and POS annotations generated by ABSynth.

\begin{figure}[ht]
\centering
\begin{subfigure}[t]{0.48\textwidth}
  \centering
  \includegraphics[width=\linewidth]{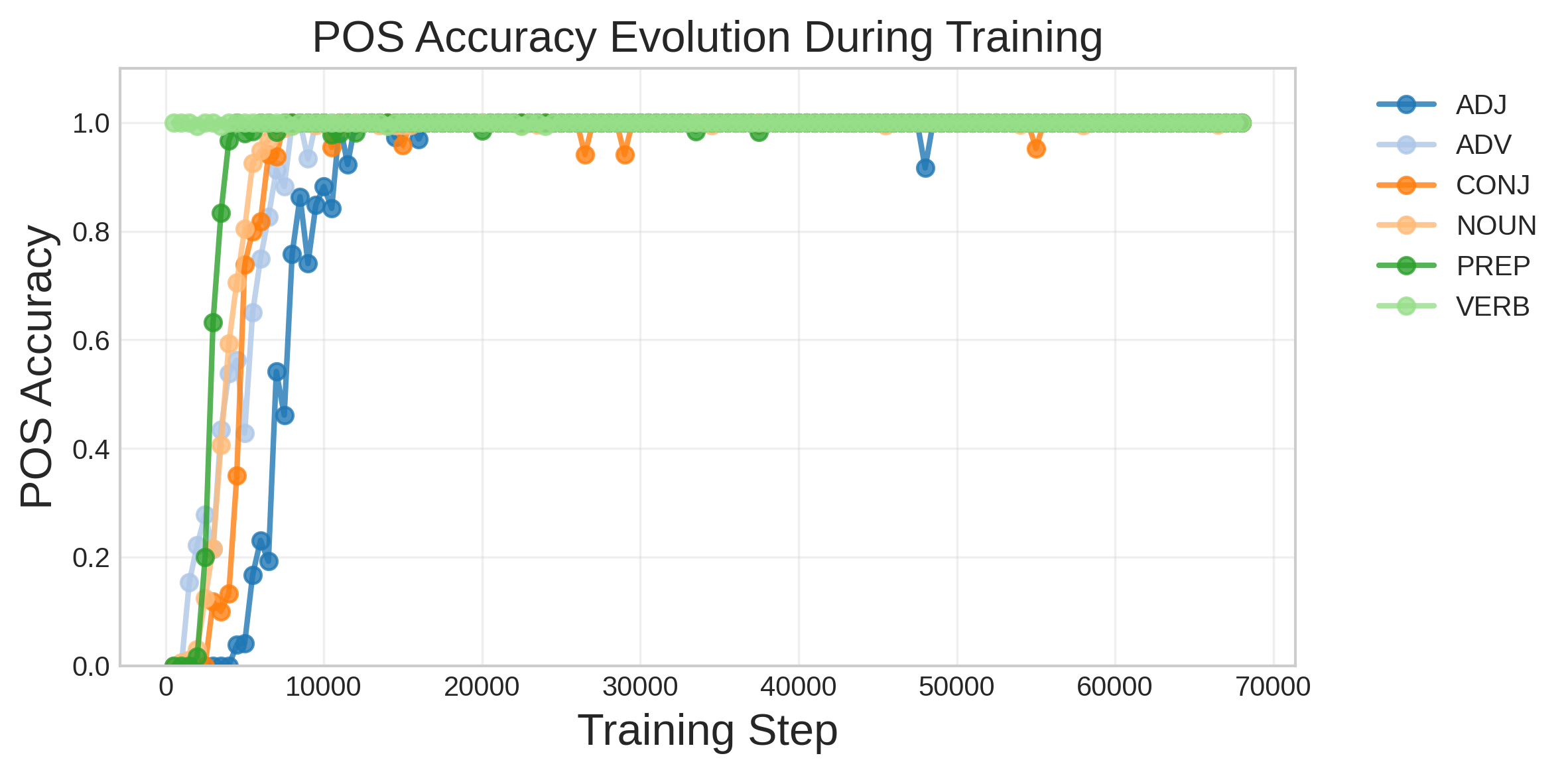}
  \caption{POS tagging accuracy across training.}
  \label{fig:pos_accuracy}
\end{subfigure}
\hfill
\begin{subfigure}[t]{0.48\textwidth}
  \centering
  \includegraphics[width=\linewidth]{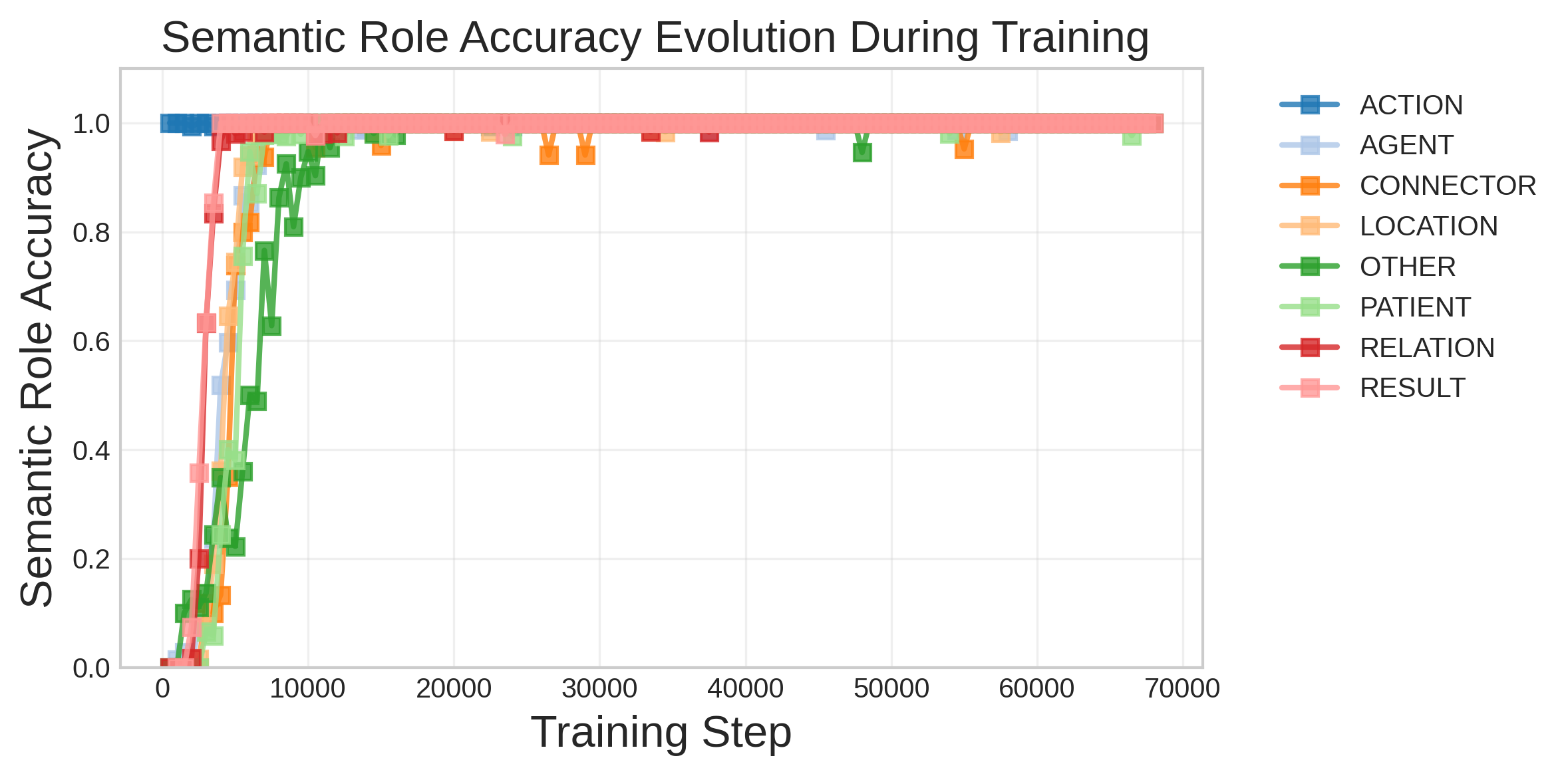}
  \caption{Semantic role prediction accuracy across training.}
  \label{fig:sr_accuracy}
\end{subfigure}
\caption{(Left) POS accuracy across training steps. (Right) Semantic role prediction accuracy.}
\label{fig:main_acc_dynamics}
\end{figure}

\paragraph{Output Example: }
Progressive analysis of model predictions across POS and semantic role categories reveals distinct learning trajectories for different linguistic abstractions, different than earlier illustrated differences. Figure~\ref{fig:main_acc_dynamics} demonstrates category-specific performance evolution, with concrete grammatical categories achieving rapid convergence while abstract semantic roles exhibit more gradual acquisition patterns.


\paragraph{Interpretation:}
\begin{itemize}

    \item \textbf{Semantic Role Acquisition Spectrum:} Roles such as \texttt{ACTION}, \texttt{RELATION}, and \texttt{RESULTS} are acquired early and maintain stable, high accuracy throughout training, indicating reliable identification of core predicate-argument relations. In contrast, roles like \texttt{LOCATION} and \texttt{PATIENT} converge more slowly and exhibit very few lingering fluctuations, reflecting their higher contextual variability and integration complexity. Notably, once core roles are learned (10K steps), the model rarely misclassifies them, even during intermediate representational reorganisations, suggesting firm grammatical competence, even if confidence (as seen in probing) continues to be refined.

    \item \textbf{POS Category Learning Hierarchy:} Among POS tags, \texttt{PREPOSITION} reaches high accuracy early, likely due to its limited variability and syntactic rigidity. \texttt{VERB} predictions stabilise quickly, note that the plotted category captures base forms rather than diverse verb types but that is also available in the tool. \texttt{NOUN} accuracy improves more gradually, reflecting greater lexical variability, but still outpaces more abstract modifiers like \texttt{ADJECTIVE} and \texttt{ADVERB}, which remain the most challenging throughout training.

    \item \textbf{Stable Predictions vs. Ongoing Representation Refinement:} Once acquired, role predictions remain stable throughout training and exhibit few regressions. This contrasts with the confidence patterns observed in probing layers, where internal representation alignment continues to shift, suggesting external outputs stabilise before internal semantics fully consolidate, which highlights a gap between external performance and internal semantic stability.

\end{itemize}

\subsection{Summary}
This walkthrough illustrates TRACE's ability to surface rich interpretability signals across diverse analytical lenses. By combining probing, dimensionality tracking, loss landscape analysis, and output performance monitoring, TRACE enables multi-faceted insight into model development over time. In this controlled ABSynth setting, TRACE reveals coherent trends in linguistic acquisition, such as early emergence of core roles, delayed abstraction in modifiers, and coordinated dynamics across representation, curvature, and output quality. Notably, we observe a divergence between surface-level grammatical accuracy and internal representational confidence, highlighting that models may produce correct outputs even as their internal abstractions continue to evolve. Such findings underscore the need for tools like TRACE to go beyond external metrics and expose latent dynamics during learning. These capabilities generalise across architectures and datasets, making TRACE a practical framework for probing both model behaviour and training dynamics in applied or research contexts.

\end{document}